\documentclass[10pt,twocolumn,letterpaper]{article}

\usepackage{iccv}
\usepackage{epsfig}

\usepackage[pagebackref=true,breaklinks=true,letterpaper=true,colorlinks,bookmarks=false]{hyperref}

\iccvfinalcopy 

\def\iccvPaperID{6372} 

\ificcvfinal\fi
\usepackage[accsupp]{axessibility} 
\usepackage{graphicx}
\usepackage{amsmath}
\usepackage{amssymb}
\usepackage{times}
\usepackage{url}            %
\usepackage{booktabs}       %
\usepackage{amsfonts}       %
\usepackage{nicefrac}       %
\usepackage{microtype}      %
\usepackage{mkolar_definitions}
\usepackage{xspace}
\usepackage{multirow}
\usepackage{algorithmic}
\usepackage{color, colortbl}
\usepackage{enumitem}
\usepackage{multirow}
\usepackage{comment}
\usepackage{bm}
\usepackage{algorithm}

\usepackage[capitalize]{cleveref}
\crefname{section}{Sec.}{Secs.}
\Crefname{section}{Section}{Sections}
\Crefname{table}{Table}{Tables}
\crefname{table}{Tab.}{Tabs.}

\begin{document}

\title{$BT^2$: Backward-compatible Training with Basis Transformation}

\author{Yifei Zhou$^*$\\
University of California, Berkeley\\
{\tt\small yifei\_zhou@berkeley.edu}
\and
Zilu Li$^*$\\
Cornell University\\
{\tt\small zl327@cornell.edu}
\and
Abhinav Shrivastava\\
University of Maryland, College Park\\
{\tt\small abhinav@cs.umd.edu}
\and
Hengshuang Zhao\\
University of Hong Kong\\
{\tt\small hszhao@cs.hku.hk}
\and
Antonio Torralba\\
MIT\\
{\tt\small torralba@mit.edu}
\and
Taipeng Tian\\
Meta AI\\
{\tt\small ttp@fb.com}
\and
Ser-Nam Lim\\
Meta AI\\
{\tt\small sernamlim@meta.com}
}

\maketitle
\def\thefootnote{*}\footnotetext{Equal contribution}
\newcommand{\brank}{R_B} \newcommand{\mrank}{R_M}
\newcommand{\olive}{\textsc{Olive}}
\newcommand{\MLE}{\mathrm{MLE}}
\newcommand{\Bayes}{\text{Bayes}}
\newcommand{\masa}[1]{\noindent{\textcolor{purple}{\{{\bf Masa:} \em #1\}}}}
\newcommand{\ming}[1]{\noindent{\textcolor{ProcessBlue}{\{{\bf Ming:} \em #1\}}}}
\newcommand{\newedit}{\color{blue}}

\begin{abstract}
Modern retrieval system often requires recomputing the representation of every piece of data in the gallery when updating to a better representation model. This process is known as backfilling and can be especially costly in the real world where the gallery often contains billions of samples. Recently, researchers have proposed the idea of Backward Compatible Training (BCT) where the new representation model can be trained with an auxiliary loss to make it backward compatible with the old representation. In this way, the new representation can be directly compared with the old representation, in principle avoiding the need for any backfilling. However, follow-up work shows that there is an inherent trade-off where a backward compatible representation model cannot simultaneously maintain the performance of the new model itself. This paper reports our ``not-so-surprising'' finding that adding extra dimensions to the representation can help here. However, we also found that naively increasing the dimension of the representation did not work. To deal with this, we propose \underline{B}ackward-compatible \underline{T}raining with a novel \underline{B}asis \underline{T}ransformation ($BT^2$). A basis transformation (BT) is basically a learnable set of parameters that applies an orthonormal transformation. Such a transformation possesses an important property whereby the original information contained in its input is retained in its output. We show in this paper how a BT can be utilized to add only the necessary amount of additional dimensions. We empirically verify the advantage of $BT^2$ over other state-of-the-art methods in a wide range of settings. We then further extend $BT^2$ to other challenging yet more practical settings, including significant changes in model architecture (CNN to Transformers), modality change, and even a series of updates in the model architecture mimicking the evolution of deep learning models in the past decade. 
Our code is available at \href{https://github.com/YifeiZhou02/BT-2}{https://github.com/YifeiZhou02/BT-2}.
\end{abstract}

\section{Introduction}
\label{sec:intro}

\begin{figure}[!ht]
  \centering
   \includegraphics[width=\linewidth]{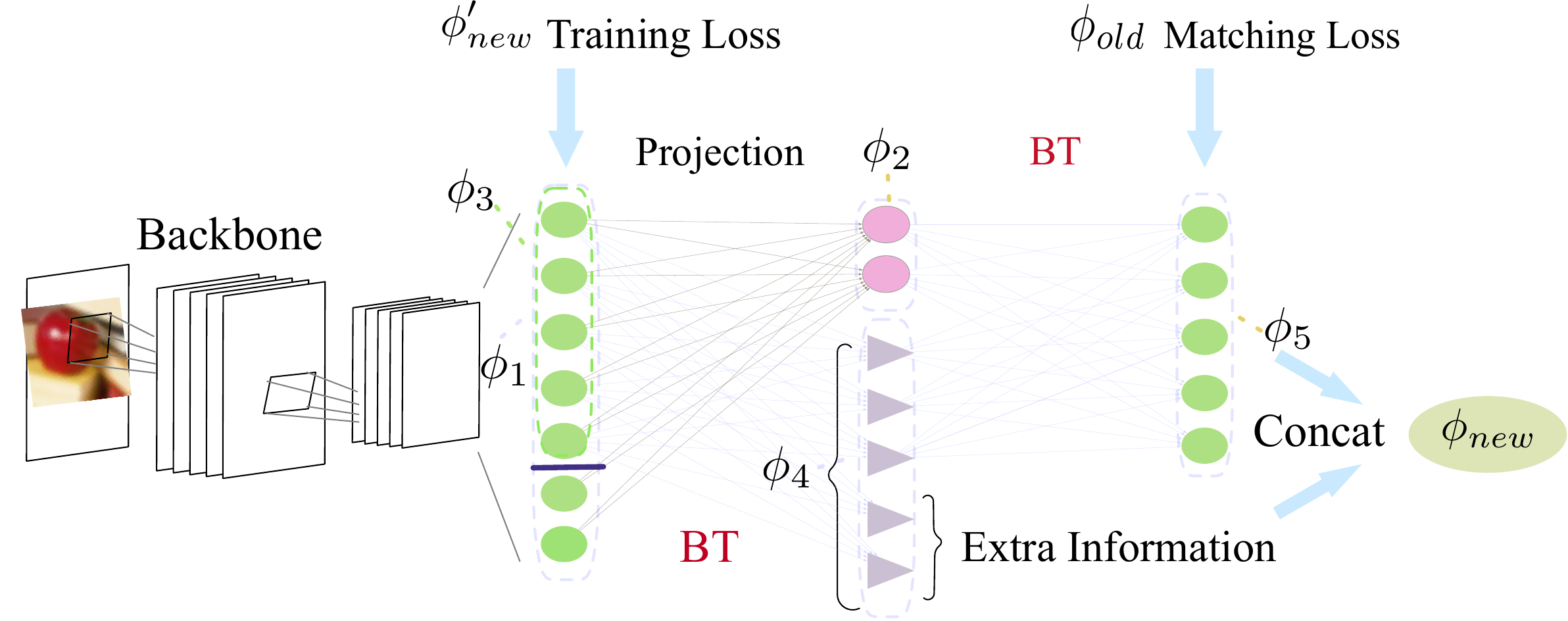}
   \caption{\textbf{Illustration of $BT^2$} The backbone produces a representation (light green ovals) that is encouraged to match the new model's representation, $\phi_{new}'$, via a matching and classification loss. A subset of this then goes through a BT transformation, which \emph{retains the information} (purple triangles) from the new representation. At the same time, the new representation is then projected into a layer (pink ovals) which is combined with part of the BT-transformed new representation (the three purple triangles). This layer then goes through a BT transformation that is then encouraged to match the old model's representation, $\phi_{old}$, in effect, resulting in a backward compatible representation as the BT transformations have to inherently retain information from both $\phi_{new}'$ and $\phi_{old}$. This is akin to the BCT procedure. The two purple triangles (i.e., what we referred to as the additional dimensions) that are not part of this are used to capture extra information in the new representation that may not be compatible. The resulting $\phi_{new}$ is then the representation used for all subsequent queries and new gallery samples. Refer to Section~\ref{sec:merge} on the definitions of $\phi_{1,2,3,4,5}$.}
   \label{fig:dimension_reduction}
\end{figure}

Modern visual retrieval systems retrieve similar images from a pool of stored data (referred to as gallery) with a given image (referred to as query). This is often done by training a model to encode all the images in the gallery and storing the resulting representations. A given query is encoded with the same model and its representation is used to retrieve the images with the most similar representations from the gallery. As better representation model design becomes available, practitioners often desire to update the representations in the gallery with the new model to achieve better performances. The issue is that if the new model has been trained independently from the old model, the representations generated by the new model will not be compatible with those generated by the old model, which necessitates re-calculating the representations of the gallery set, a process known as ``backfilling'' \cite{9156906}. This process gets very costly or even impossible for real world galleries which often contain billions and billions of images.

Shen \etal \cite{9156906} therefore proposes a framework to train the new model while being compatible with the old model, known as Backward Compatible Training (BCT), with the hope of removing the need for backfilling. They propose to add an ``influence loss'' to the training objective of the new model to heuristically induce a backward-compatible representation. 
However, as pointed out by \cite{ramanujan2022forward}, adding this influence loss can significantly hurt the performance of the new model when compared to its independently trained counterpart. To mitigate this issue, subsequent works \cite{DBLP:journals/corr/abs-2201-09724, DBLP:journals/corr/abs-2108-01958, https://doi.org/10.48550/arxiv.2203.01583} have proposed various more sophisticated influence losses, but these endeavors have achieved limited success. Indeed, as shall be detailed in Section~\ref{sec:tradeoff}, it may be impossible to find a new representation model that is at the same time backward compatible yet achieves the fullest potential of the new model. In view of this, another line of work in which researchers utilize a light-weight transformation of the old representation into the new representation \cite{DBLP:journals/corr/abs-2008-04821, ramanujan2022forward, https://doi.org/10.48550/arxiv.2204.13919} looks promising. However, despite their effort to make the transformation light-weight, it still requires a costly procedure of applying a neural network to update billions of images in the gallery. 

In this paper, we present findings that the conflict between backward compatibility and new model performance can be mitigated by expanding the representation space to simultaneously accommodate both the old model and the best independently trained new model. To motivate this, one can first consider an upper bound solution along this direction, where the representation of the old model is concatenated with that of an independently trained new model - being independently trained, the new model is no more limited by the backward compatibility requirement. Subsequently, queries and new samples added to the gallery are now encoded with the concatenated representations. During retrieval, since it is easy to distinguish between the gallery samples that are still of the old representations and those with the concatenated representations due to the difference in size, we can simply truncate the new representation from the query when comparing with the old representations in the gallery. This upper bound solution is ``perfectly'' backward compatible but suffers from two critical drawbacks: (1) it significantly increases computations due to the additional number of forward passes when computing the query representation, and (2) it begets a significant dimension expansion as a result of the concatenation. In fact, both (1) and (2) can get especially severe after multiple model updates. 

Nevertheless, such an upper bound solution provides us with the inspiration to consider adding dimensions to the representation as \emph{necessary} while conducting BCT. We first tried naively adding dimensions (e.g., directly adding an extra 32 dimensions while training a BCT model) to the new representation, but found that this did not lead to a clear advantage as shown in Section \ref{sec:experiments}. Instead, we conjecture and show that what would be more desirable is to add dimensions for the purpose of storing any information that is not compatible between the old and new representation. Towards this end, we propose a novel \underline{B}ackward-compatible \underline{T}raining with \underline{B}asis \underline{T}ransformation ($BT^2$) that exploits a series of learnable basis transformations (BT) to find the information in the new representation that is incompatible with the old representation. Because a BT is basically an orthonormal transformation, the output of a BT retains the entirety of the information stored in the input (see Lemma~\ref{lemma:change of basis}). With this in mind, we introduce some clever manipulation with BT that helps to exactly ``force'' incompatible information in the new representation into the additional dimensions, while keeping the compatible information in the BCT representation. Fig.~\ref{fig:dimension_reduction} provides a conceptual explanation of our $BT^2$ design.

In summary, our contributions are three-fold:
\begin{itemize}[itemsep=0mm, topsep=2mm]
    \item We show that the dilemma between backward compatibility and new model development can be reconciled with extra dimensions.
    
    \item We propose $BT^2$ that exploits a series of learnable changes of basis to effectively exploit the extra dimensions, and verify its empirical advantage over other state-of-the-art methods in a wide range of backward compatibility tasks.
    
    \item We extend $BT^2$ to more challenging and practical scenarios that have not been considered by existing works to the best of our knowledge. These include significant changes in model architecture, compatibility between different modalities, and even a series of updates in the model architecture mimicking the history of deep learning in the past decade. 
\end{itemize}
\section{Related Works}
\paragraph{Model Compatibility and Backward Compatibility.}
Model compatibility has received an increasing amount of attention in the research community due to its practical utility \cite{DBLP:journals/corr/abs-2108-01958, DBLP:journals/corr/abs-2008-04821, 8953998, https://doi.org/10.48550/arxiv.2107.01057}, where the goal is to learn a shared representation space in which representations from different models can be directly compared. In particular, backward compatibility was introduced in \cite{9156906}, where the authors proposed an influence loss that tries to move the new and old representations closer. Subsequent works either introduce a transformation module \cite{Hu_2022, ramanujan2022forward, DBLP:journals/corr/abs-2008-04821} or enhanced regularization loss functions \cite{DBLP:journals/corr/abs-2108-01958, DBLP:journals/corr/abs-2201-09724, https://doi.org/10.48550/arxiv.2203.01583}. However, some key disadvantages associated with these approaches include that some of them depend on an auxilliary loss that prevents the new model from reaching its fullest potential, while others still require a lightweight backfilling. For the latter, a recent work known as Forward Compatible Training (FCT) \cite{ramanujan2022forward} has been proposed that trains a lightweight transformation module to transform the old representations into new representations for backward compatibility. However, \emph{a key difference between this paper with FCT is that FCT still requires lightweight backfilling and a side-information model (which hopefully contains sufficient information to train the transformation module) but those are not required in this paper}. 


\paragraph{Continual Learning and Transfer Learning.}
The field of backward compatible representation learning is also related to continual learning \cite{Delange_2021, https://doi.org/10.48550/arxiv.1611.06194, DBLP:journals/corr/abs-1812-00420, DBLP:journals/corr/RebuffiKL16, DBLP:journals/corr/abs-1905-11614} and transfer learning \cite{DBLP:journals/corr/abs-1911-02685, DBLP:journals/corr/ZhangY17aa, 5288526, Zhao2010OTLAF, LU201514}. However, these two fields have different focuses. Continual learning focuses on training a model to perform well on a new task without forgetting the old task, and transfer learning focus on transferring a model to perform well on a different domain with the original training domain. On the other hand, backward compatible representation learning focuses on the same task, i.e., representation learning, such that the representation from the improved new model can be directly used to compare with the old model. 
\section{Methodology}
To facilitate discussions, we follow standard notations in this field, using $\phi_1/\phi_2$ to represent the retrieval performance of using $\phi_1$ for representing queries while using $\phi_2$ for representing samples in the gallery. We also denote the representations from the old and final new model as $\phi_{old}$ and $\phi_{new}$ respectively. $\phi_{new}'$ is the independently trained new model representation while $\phi_{new}$ is the final new model representation after combining $\phi_{old}$ and $\phi_{new}'$. Our goal is to learn the best $\phi_{new}$ possible (either when more data and/or better model architectures becomes available) while respecting the following commonly accepted criterion for backward compatible representations \cite{9156906, DBLP:journals/corr/abs-2008-04821}.

\subsection{Problem Setup}
\label{sec:problem setup}

\begin{definition}[Backward Compatibility]
{ $\phi_{new}$ is backward compatible with $\phi_{old}$, if $\forall x_i, x_j$ from the distribution of interest,
\begin{align*} 
    d(\phi_{new}(x_i), \phi_{old}(x_j)) \ge d(\phi_{old}(x_i), \phi_{old}(x_j)), \forall y_i \ne y_j,\\ 
    d(\phi_{new}(x_i), \phi_{old}(x_j)) \le d(\phi_{old}(x_i), \phi_{old}(x_j)), \forall y_i = y_j,
\end{align*}
where d is a distance measure, and $y_i, y_j$ are the corresponding labels of $x_i, x_j$.
\label{backward compatibility}
}
\end{definition}

\noindent Alternatively, to relax the above criterion that enforce the constraints on every data point, if there is some empirical metrics $M(\phi_1, \phi_2)$ (for example, top1 accuracy for $\phi_1/\phi_2$), we consider backward compatibility as $M(\phi_{new}, \phi_{old}) >= M(\phi_{old}, \phi_{old})$. In addition, we desired another key property when learning $\phi_{new}$:
 
\begin{definition}[Not Hurting New Model] \label{no hurt}
 $\phi_{new}$ is said to be not hurting the new model if, $\forall x_i, x_j$ from the distribution of interest:
 \begin{align*}
    d(\phi_{new}(x_i), \phi_{new}(x_j)) = d(\phi_{new}'(x_i), \phi_{new}'(x_j)).
\end{align*}
\end{definition}

\noindent Similarly, if we relax it with an empirical metric $M$, the definition becomes $M(\phi_{new}, \phi_{new}) >= M(\phi_{new}', \phi_{new}')$. In this paper, we adopt the the negative dot product as distance metric for simplicity:
\begin{align*}
    d(a,b) = - a^\top b
\end{align*}
Note that in this paper, all final representations are normalized, so this dot product is equivalent to cosine similarity up to a constant multiplier.

\subsection{Backward Compatibility vs $\phi_{new}$ Performance}
\label{sec:tradeoff}

We argue that only adding an influence loss \cite{9156906, DBLP:journals/corr/abs-2201-09724, DBLP:journals/corr/abs-2108-01958} fails to reliably guarantee backward compatibility without hurting the new model, and this idea is formalized in Lemma \ref{lemma:counterexample}. The implication of this Lemma \ref{lemma:counterexample} can be considered as an inherent trade-off between backward compatibility and $\phi_{new}$ performance, which inspires us to ``hold'' incompatible information of the new model on the additional orthogonal dimensions to avoid this conflict.


\begin{lemma} \label{lemma:counterexample}
There exist cases where backward compatibility will significantly limit the potential of the new model while using negative cosine similarity as the distance metric.
\end{lemma}


\subsection{Learnable Basis Transformation}
\label{sec:change of basis}
We make heavy use of basis transformation (BT). A BT
is essentially a learnable change of basis represented by an orthonormal matrix, $P$. $P$ can be parameterized as the exponential of a left skew-symmetric $A$, so that $P = e^A$, where the upper entries in A are learnable parameters. This design is made possible by the following Lemma~\ref{lemma:matrix exponentiation}.

\begin{lemma}
\label{lemma:matrix exponentiation}
    If $A$ is a left skew-symmetric matrix, $P = e^A$ is orthonormal where $P = e^A$ is defined as:
    \begin{align*}
        P = e^A = \sum_{k = 0}^{\infty} \frac{A^k}{k!}
    \end{align*}
\end{lemma}

\noindent Intuitively, for any representation $\phi(x)$, applying a change of basis on $\phi(x)$ to get $P\phi(x)$ should not lose any information and will not hurt the quality of the $\phi$. This intuition can be formalized by the following Lemma~\ref{lemma:change of basis}.

\begin{lemma}
\label{lemma:change of basis}
    If $P$ is an orthonormal matrix of dimension $m \times m$, $\forall \phi(x_1), \phi(x_2)$ that are $m$ dimensional vectors:
    \begin{align*}
        \phi(x_1)^\top \phi(x_2) = (P\phi(x_1))^\top (P(\phi(x_2))
    \end{align*}
\end{lemma}

\subsection{Merging $\phi_{new}'$ and $\phi_{old}$ with BT}
\label{sec:merge}

With all the premises set up in the previous sections, we will now describe our proposal on achieving both criteria in definition~\ref{backward compatibility} and \ref{no hurt}. Further, we would like to to minimize the additional dimensions required to do that (Lemma~\ref{lemma:counterexample}) as well as ensuring that $\phi_{new}$ requires only a single forward pass to obtain. Our proposal utilizes the BT (Lemma~\ref{lemma:matrix exponentiation}) as described in Fig.~\ref{fig:dimension_reduction}. We conjecture that although new and old models might differ in their model architecture or training data, they should still encode a lot of information in common. Therefore, we propose to train $\phi_{new}$ to automatically pick out the information from $\phi_{new}'$ not compatible with $\phi_{old}$, and only encode this extra information on the dimension orthogonal to $\phi_{old}$. This can be realized by restricting ourselves to using only learnable change of basis, thanks to the information-preserving property detailed in Lemma \ref{lemma:change of basis}.

\begin{algorithm}[t]
\begin{algorithmic} 
\caption{Dimension Reduction by Learnable Change of Basis}
\label{learnable basis algorithm}
\REQUIRE Learnable backbone $F$ with output dimension $m+n$, learnable projection layer $f$ mapping from dimension $m+n$ to $d$, BT block $B_1$ of $m \times m$ and $B_2$ of $n \times n$. Constant C and image x.
\STATE $\phi_1 \gets F(x)$
\STATE $\phi_2 \gets f(\phi_1)$, $\phi_2 \gets \frac{\phi_2}{\|\phi_2\|}$
\STATE $\phi_3 \gets \frac{\phi_1[:m]}{\|\phi_1[:m]\|}$
\STATE $\phi_4 \gets C B_1 \phi_3$
\STATE $\phi_5 \gets B_2 \begin{bmatrix} \begin{matrix} &\phi_2 \\ &\phi_4[:n-d]
\end{matrix} \end{bmatrix}$
\STATE $\phi_{new} \gets \begin{bmatrix} \begin{matrix} &\phi_5 \\ &\phi_4[n-d:]
\end{matrix} \end{bmatrix}$
\end{algorithmic}
\end{algorithm}

Referring to Fig.~\ref{fig:dimension_reduction}, we mainly add two BT layers. Concretely, suppose $\phi_{new}', \phi_{old}$ are of dimension $m,n$ respectively, and we allow a dimension expansion of $d$, so that $\phi_{new}$ has a dimension of $m+d$ in our budget. Our backbone produces $\phi_1$ of dimension $m+n$, and the first $m$ dimension (referred to as $\phi_3$) is trained to mimic $\phi_{new}'$ with the same loss that is used to train $\phi_{new}'$. Notably, there is no backward influence loss on $\phi_3$ so that it is hoped $\phi_3$ can be as good as independent $\phi_{new}'$. 

After that, we pass the entire $\phi_1$ into the projection layer to produce a $\phi_2$ of dimension $d$ which will be some additional features of $\phi_{old}$ that are not used by $\phi_{new}'$. $\phi_3$ is passed into the first BT layer $B_1$ to get $\phi_4$, allowing us to split $\phi_3$ into compatible information with the old representation (the first $n-d$ dimensions) and incompatible information (the remaining $m - n + d$ dimensions). The first $n-d$ dimension is concatenated with $\phi_2$ and passed into the second BT layer $B_2$ to mix the information of $\phi_2$ and from $\phi_4$ to get $\phi_5$. Similarly, $\phi_5$ is trained to mimic $\phi_{old}$ and there is no new model training loss so it is hoped that $\phi_5$ can be as compatible to the old representations as possible. Lastly, $\phi_5$ is concatenated with the remaining $m-n+d$ dimension of $\phi_4$ to get the final $\phi_{new}$ of dimension $m+d$. Basis transformations $B_1$ and $B_2$ are designed such that all the information is preserved from $\phi_3$ that is trained to match an independent model $\phi_{new}'$. Algorithm \ref{learnable basis algorithm} detailed our proposed method, with additional normalization details to ensure the information of $\phi_{new}'$ dominates the information of $\phi_{old}$ in the final representation $\phi_{new}$ via the factor $C$. Note that this does not cause $\phi_{new}/\phi_{old}$ to suffer since we truncate $\phi_{new}$ when comparing to $\phi_{old}$, effectively eliminating the extra incompatible information between $\phi_{new}'$ and $\phi_{old}$.



Intuitively, the BT layers serve the purpose of retaining the entirety of the information between the input and the output, which as as result means that $\phi_5$ is akin to a BCT procedure. It also means that any incompatible information between $\phi_{new}'$ and $\phi_{old}$ is ``forced'' into $\phi_4[n-d:]$ as a result of the training.
\section{Experiments}
\label{sec:experiments}
We provide experimental results that (1) benchmark our method with existing backward compatible representation learning methods based on both criteria in definition~\ref{backward compatibility} and \ref{no hurt}, (2) demonstrate our method's ability to handle cases such as data changes or what \cite{9156906} referred to as open classes (\eg the old model is trained on 500 Imagenet classes while the new model on 1000 Imagenet classes, with both using the same architecture), significant changes in model architecture (e.g., ResNet to pretrained Transformers), different modalities, and a series changes of model architecture (mimicking the historical development of deep learning), and (3) ablating the effect of the number of extra dimensions on the performance of our method.

\subsection{Datasets}
This work makes use of the following datasets:
\begin{itemize}
    \item \textbf{Cifar-100}\cite{cifar100}: Cifar-100 is a popular image classification dataset consisting of 60000 images in 100 classes. We use Cifar-50 to refer to the partition consisting of all the images from the first 50 classes.
    
    \item \textbf{Imagenet-1k} \cite{5206848}: Imagenet-1k is a large-scale image recognition dataset for ILSVRC 2012 challenge. It has 1000 image classes with approximately 1.2k images per class. We follow the same partition as \cite{ramanujan2022forward} where we use the images from the first 500 classes as Imagenet-500. \footnote{We use the class split from https://gist.github.com/aaronpolhamus-
    /964a4411c0906315deb9f4a3723aac57 .}
\end{itemize}

\subsection{Evaluation Metrics}

\noindent \textbf{Cumulative Matching Characteristics (CMC).} CMC corresponds to the top-k accuracy, where we sort the gallery representations by their similarity to the query representation. It is considered correct if a match with the same class is in the first k gallery representations. We report CMC top-1 and top-5 for all models. 

\noindent \textbf{Mean Average Precision (mAP).} mAP is a standard metrics that summarizes precision and recall metrics by taking the area under the precision-recall curve. We compute the average precision in the recall range [0.0, 1.0].

\subsection{Baselines}
We benchmark against the following baselines to validate our method. 

\noindent \textbf{Independent.} For this baseline, it is basically the two models $\phi_{new}'$ and $\phi_{old}$ without accounting for any backward compatibility. $\phi_{old}/\phi_{old}$ and $\phi_{new}'/\phi_{old}$ respectively provide a rough upper and lower bound for $\phi_{new}/\phi_{old}$, while $\phi_{new}'/\phi_{new}'$ provides a rough upper bound for $\phi_{new}/\phi_{new}$.


\noindent \textbf{Backward Compatible Training (BCT).} BCT was introduced in Shen \etal \cite{9156906}. As the first attempt for backward compatibility, it is frequently adopted as the baseline in many recent papers\cite{ramanujan2022forward, https://doi.org/10.48550/arxiv.2204.13919, Hu_2022}. Specifically, BCT utilizes a classification loss but adds an ``influence loss'' during training to achieve backward compatibility. Denoting $w_{\phi}$ as the new representation model, $w^c$ a trainable classification head, and $w^c_{old}$ the fixed classification head that was trained with the old representation head, the following loss terms are used in BCT:
\begin{align*}
    \mathcal{L}_{BCT}(\phi, w^c, x) = \mathcal{L}(w^c, \phi|x) + \lambda \mathcal{L}(w^c_{old}, \phi|x),
\end{align*}
where $\lambda$ is a hyperparameter to tune, and both $\mathcal{L}(w^c, \phi|x)$ and $\mathcal{L}(w^c_{old}, \phi|x)$ represent a classification loss with $w^c$/$w^c_{old}$ as the classification head and $\phi$ as the representation. Backward propagation trains $\phi$ and $w_{\phi}$.

\noindent \textbf{BCT with 32 Extra Dimensions (BCT(+32)).} To ensure a fair comparison with our method, where we add 32 dimensions, we create a variant of BCT with the dimension expanded by 32. We use the same loss function as BCT except that we pad the missing dimension of $w^c_{old}$ with 0.

\noindent \textbf{Regression-alleviating Compatibility Regularization (Contrast).} This is a recently proposed method with a more sophisticated auxiliary loss \cite{DBLP:journals/corr/abs-2201-09724} to replace influence loss:
\begin{align*}
    &\mathcal{L}_{ra-comp}(\phi_{new}, x) =\\
    & \log (1 + \frac{\sum_{k \in \mathcal{B} - p(x)}\exp{\phi_{new}(k) \cdot \phi_{old}(k)/\tau}}{\exp{\phi_{new}(x) \cdot \phi_{old}(x)/\tau}}\\
    +& \frac{\sum_{k \in \mathcal{B} - p(x)} \exp{\phi_{new}(k) \cdot \phi_{new}(k)/\tau}}{\exp{\phi_{new}(x) \cdot \phi_{old}(x)/\tau}}),
\end{align*}
where $\mathcal{B}$ is the mini-batch for training, $p(x)$ is the set of samples in the minibatch with the same label as $x$, and $\tau$ is a temperature hyperparameter.

\noindent \textbf{$BT^2$ (Ours).} We use a dimension expansion of 32, and a classification loss that is the same as that used to train $\phi_{new}'$, and a combination of cosine similarity loss and BCT influence loss for matching loss for $\phi_{old}$. Specifically, we use the following loss as ``$\phi_{new}'$ training loss'' for $\phi_3$:
\begin{align*}
    \mathcal{L}_{\phi_{new}'} (\phi_3, w^c_{\phi_3}, x) = \mathcal{L}(w^c_{\phi_3}, \phi_3|x) + \lambda_1(1 - \phi_3^\top \phi_{new}'),
\end{align*}
where $\lambda_1$ is a hyperparameter and $\phi_{new}'$ is an independently trained model, both $\phi_3$ and $w^c_{\phi_3}$ are trained. Similarly, we use the following loss as ``$\phi_{old}$ matching loss'':
\begin{align*}
    \mathcal{L}_{\phi_{old}} (\phi_5, x) = \lambda_2 \mathcal{L}(w^c_{old}, \phi_5|x) + \lambda_3(1 - \phi_5^\top \phi_{old}),
\end{align*}
where $\lambda_2, \lambda_3$ are two hyperparameters and $w^c_{old}$ is the fixed classification head used to train $\phi_{old}$. C is taken to be 2. For transformer-based models, we found it helpful to apply the classification loss to the final representation $\phi_{new}$ instead of $\phi_{3}$.

\subsection{Implementations Details}
\noindent \textbf{Experiments on Table 1, 2, 3.} Experiments for these tables are carried out on 8x Nvidia 2080Ti. For all baselines and methods, transformer models are finetuned with sgd optimizer with a learning rate 0.01 and batch size 64 for 25 epochs, while ResNet50 models are trained with adam optimizer with a learning rate 0.001 and batch size 256 or 128 for 100 epochs.

\noindent \textbf{Experiments on Table 4, 5, 6, 7.} Experiments for these tables are carried out on 8x Nvidia A100. For all baselines and methods, transformer models are finetuned with sgd optimizer with a learning rate 0.01 and batch size 512 for 25 epochs, while AlexNet, ResNet50 models are trained with adam optimizer with a learning rate 0.001 and batch size 2048 for 100 epochs. VGGNet-13 with batch normalization is trained with adam optimizer with a learning rate 0.001 and batch size 1024 for 100 epochs.

\subsection{Data Change}

\begin{table}
\begin{center}
\resizebox{\columnwidth}{!}{%
\begin{tabular}{ c c c c }
\toprule[1pt]
\multirow{2}{5em}{Method} & \multirow{2}{3em}{Case} & CMC & \multirow{2}{5em}{mAP@1.0} \\
& & top1-top5 &\\
\hline
\multirow{3}{5em}{Independent} & $\phi_{old}/\phi_{old}$ & 33.6-55.4 & 24.4\\
& $\phi_{new}'/\phi_{old}$ & 0.8-4.9 & 1.5\\
& $\phi_{new}'/\phi_{new}'$ & 62.7-74.6 & 49.9\\
\hline
\multirow{2}{5em}{BCT} & $\phi_{new}^{bct}/\phi_{old}$  & 23.5-60.4 & 23.9\\
& $\phi_{new}^{bct}/\phi_{new}^{bct}$ & 56.1-70.8 & 43.6\\
\hline
\multirow{2}{5em}{BCT (+32)} & $\phi_{new}^{bct(+32)}/\phi_{old}$ & 22.1-61.3& 23.8\\
& $\phi_{new}^{bct(+32)}/\phi_{new}^{bct(+32)}$ & 56.1-71.3 & 44.1\\
\hline
\multirow{2}{5em}{Contrast} & $\phi_{new}^{contrast}/\phi_{old}$ & 26.1-61.8 & 25.1\\
& $\phi_{new}^{contrast}/\phi_{new}^{contrast}$ & 57.9-75.2 & 36.7\\
\hline
\multirow{2}{5em}{$BT^2$ (Ours)} & $\phi_{new}^{bt^2}/\phi_{old}$ & \textbf{38.7}-\textbf{67.1} & \textbf{28.0}\\
& $\phi_{new}^{bt^2}/\phi_{new}^{bt^2}$ & \textbf{64.4}-\textbf{78.7} & \textbf{53.2}\\
\bottomrule[1pt]
\end{tabular}}
\caption{Backward compatible experiments on Cifar-50 to Cifar-100 with only data change. Both the old model and the new model uses Resnet50-128 architecure.}
\label{table:Cifar-100 data change}
\end{center}
\vspace{-5mm}
\end{table}

\begin{table}
\begin{center}
\resizebox{\columnwidth}{!}{%
\begin{tabular}{ c c c c }
\toprule[1pt]
\multirow{2}{5em}{Method} & \multirow{2}{3em}{Case} & CMC & \multirow{2}{5em}{mAP@1.0} \\
& & top1-top5 &\\
\hline
\multirow{3}{5em}{Independent} & $\phi_{old}/\phi_{old}$ & 43.1-58.3 & 30.9\\
& $\phi_{new}'/\phi_{old}$ &0.1-0.5  & 0.2\\
& $\phi_{new}'/\phi_{new}'$ & 67.9-81.4 & 52.3\\
\hline
\multirow{2}{5em}{BCT} & $\phi_{new}^{bct}/\phi_{old}$  & 41.3-64.4 & 33.3\\
& $\phi_{new}^{bct}/\phi_{new}^{bct}$ & 63.7-79.0 & 51.2\\
\hline
\multirow{2}{5em}{BCT (+32)} & $\phi_{new}^{bct(+32)}/\phi_{old}$ & 37.4-64.4 & 30.0\\
& $\phi_{new}^{bct(+32)}/\phi_{new}^{bct(+32)}$ & 65.7-80.1 & 52.0\\
\hline 
\multirow{2}{5em}{Contrast} & $\phi_{new}^{contrast}/\phi_{old}$ & 39.0-66.7 & 29.4\\
& $\phi_{new}^{Contrast}/\phi_{new}^{contrast}$ & 65.6-\textbf{81.2} & 47.6\\
\hline
\multirow{2}{5em}{$BT^2$ (Ours)} & $\phi_{new}^{bt^2}/\phi_{old}$ & \textbf{47.8}-\textbf{68.0} & \textbf{33.8}\\
& $\phi_{new}^{bt^2}/\phi_{new}^{bt^2}$ & \textbf{66.5}-80.9 & \textbf{54.4}\\
\bottomrule[1pt]
\end{tabular}}
\caption{Backward compatible experiments on Imagenet-500 to Imagenet-1k with only data change. Both the old model and the new model uses Resnet50-128 architecure.}
\label{table:Imagenet data change}
\end{center}
\vspace{-5mm}
\end{table}
\noindent \textbf{Cifar-50 to Cifar-100.} In this experiment, the model for $\phi_{old}$ is a ResNet50 with an output feature dimension of size 128, and trained on Cifar-50. The model for $\phi_{new}$ is also a ResNet50 but trained on the entire Cifar-100 dataset. For retrieval, we use the Cifar-100 validation set as both the gallery set and the query set.

\noindent \textbf{Imagenet-500 to Imagenet-1k.} Same as above except the model for $\phi_{old}$ is trained on Imagenet-500, and that for $\phi_{new}$ on the entire Imagenet-1k dataset. For retrieval, we use the Imagenet-1k validation set as both gallery and query.

\noindent \textbf{Results.} The results are shown in Table \ref{table:Cifar-100 data change} and \ref{table:Imagenet data change}, where we observe that BCT can indeed achieve backward compatibility on large-scale image classification datasets like Imagenet and also achieves reasonable performance on $\phi^{bct}_{new}/\phi^{bct}_{new}$. However, as has also been discussed in the previous literature, $\phi^{bct}_{new}/\phi^{bct}_{new}$ is significantly influenced by the auxiliary influence loss. Comparing to the upper bound of training independently, $\phi_{new}'/\phi_{new}'$, BCT is only 63.7\% and 56.1\% for CMC top 1 on Imagenet and Cifar-100 respectively. Furthermore, the backward compatibility of BCT can be unstable in some of the datasets where we observe that $\phi^{bct}_{new}/\phi_{old}$ is only 23.5\% for CMC top 1 on Cifar-100 while $\phi_{old}/\phi_{old}$ is 33.6\%. Its unstable performance might be because the influence loss in BCT does not directly encourage the model to be compatible with the old representation, but rather to be compatible with the old classification head, and the inherent conflict between $\phi^{bct}_{new}/\phi_{old}$ and $\phi^{bct}_{new}/\phi^{bct}_{new}$. We also observe a similar pattern for Contrast, indicating that even a more sophisticated influence loss may not suffice in overcoming this issue. In contrast, with the extra dimension introduced by $BT^2$, we observe a significant improvement across the board. In particular, for CMC top 1 on Imagenet, $BT^2$ achieves 47.8\% on $\phi^{bt^2}_{new}/\phi_{old}$ and 66.5\% on $\phi_{new}^{bt^2}/\phi_{new}^{bt^2}$, which are a 6.5\% and 2.8\% improvement over BCT respectively. It shows that $BT^2$ can mitigate the trade-off between backward compatibility and performance of the new model, by exploiting extra dimensions. We would like to also note that just naively adding new dimension like BCT (+32) did not show a clear improvement over BCT, which highlights the importance of adding dimensions in a principled manner.

\subsection{Model Change}
\label{sec:model change}
\noindent \textbf{Cifar-50 to Cifar-100 (ResNet50 to Transformer).} The setting of this experiment is similar to Cifar-50 to Cifar-100 except that the new model is finetuned from ``ViT-B16'' \cite{DBLP:journals/corr/abs-2010-11929, NIPS2017_3f5ee243} pretrained on entire Imagenet-21k training set. 

\noindent \textbf{Imagenet-500 to Imagenet-1k (ResNet50 to Transformer).} The setting of this experiment is similar to Imagenet-500 to Imagenet-1k except that the new model is from the same ``ViT-B16'' \cite{DBLP:journals/corr/abs-2010-11929} pretrained on entire Imagenet-21k training set. 

\noindent \textbf{Results.} The results are shown in Table \ref{table:Imagenet model change}. This is a challenging setting which, to be best of our knowledge, has not been studied in the previous literature. We observe a similar pattern to the data change setting. Designing sophisticated backward compatible loss or naively adding dimensions, as in the case of Contrast and BCT (+32), did not produce any clear improvement. However, $BT^2$ outperforms BCT by 12.5\% and 2.1 \% in terms of CMC top 1 on Imagenet $\phi_{new}^{bct}/\phi_{old}$ and $\phi_{new}^{bct}/\phi_{new}^{bct}$ respectively. 

\begin{table}
\begin{center}
\resizebox{\columnwidth}{!}{%
\begin{tabular}{ c c c c }
\toprule[1pt]
\multirow{2}{5em}{Method} & \multirow{2}{3em}{Case} & CMC & \multirow{2}{5em}{mAP@1.0} \\
& & top1-top5 &\\
\hline
\multirow{3}{5em}{Independent} & $\phi_{old}/\phi_{old}$ & 33.6-55.4 & 24.4\\
& $\phi_{new}'/\phi_{old}$ & 0.3-4.7 & 1.7\\
& $\phi_{new}'/\phi_{new}'$ & 89.5-94.3 & 87.5\\
\hline
\multirow{2}{5em}{BCT} & $\phi_{new}^{bct}/\phi_{old}$  & 45.7-83.7  & 32.9\\
& $\phi_{new}^{bct}/\phi_{new}^{bct}$ & 88.7-93.5 & 85.7\\
\hline
\multirow{2}{5em}{BCT (+32)} & $\phi_{new}^{bct(+32)}/\phi_{old}$ & 44.9-\textbf{86.2} & 32.7\\
& $\phi_{new}^{bct(+32)}/\phi_{new}^{bct(+32)}$ & 88.6-93.7 & 84.8\\
\hline
\multirow{2}{5em}{Contrast} & $\phi_{new}^{contrast}/\phi_{old}$ & 45.6-81.0 & 32.8\\
& $\phi_{new}^{contrast}/\phi_{new}^{contrast}$ & 88.2-94.0 & 81.9\\
\hline
\multirow{2}{5em}{$BT^2$ (Ours)} & $\phi_{new}^{bt^2}/\phi_{old}$ & \textbf{51.2}-85.5 & \textbf{34.0}\\
& $\phi_{new}^{bt^2}/\phi_{new}^{bt^2}$ & \textbf{90.0}-\textbf{94.8} & \textbf{88.4}\\
\bottomrule[1pt]
\end{tabular}}
\caption{Backward compatible experiments on cifar-50 to cifar-100 with both data change and model change. The old model uses Resnet50-128 architecure, while the new model uses a transformer ``ViT-B16'' \cite{DBLP:journals/corr/abs-2010-11929} pretrained on full training set of Imagenet21K.}
\label{table:Cifar-100 model change}
\end{center}
\vspace{-5mm}
\end{table}
\begin{table}
\begin{center}
\resizebox{\columnwidth}{!}{%
\begin{tabular}{ c c c c }
\toprule[1pt]
\multirow{2}{5em}{Method} & \multirow{2}{3em}{Case} & CMC & \multirow{2}{5em}{mAP@1.0} \\
& & top1-top5 &\\
\hline
\multirow{3}{5em}{Independent} & $\phi_{old}/\phi_{old}$ & 43.1-58.3 & 30.9\\
& $\phi_{new}'/\phi_{old}$ & 0.0-0.2 & 0.1\\
& $\phi_{new}'/\phi_{new}'$ & 78.0-87.5 & 72.4\\
\hline
\multirow{2}{5em}{BCT} & $\phi_{new}^{bct}/\phi_{old}$  & 41.1-69.5 & 33.0\\
& $\phi_{new}^{bct}/\phi_{new}^{bct}$ & 74.8-86.7 & 66.5\\
\hline
\multirow{2}{5em}{BCT (+32)} & $\phi_{new}^{bct(+32)}/\phi_{old}$ & 40.5-69.4 & 32.9\\
& $\phi_{new}^{bct(+32)}/\phi_{new}^{bct(+32)}$ & 75.1-87.1 & 66.8\\
\hline
\multirow{2}{5em}{Contrast} & $\phi_{new}^{contrast}/\phi_{old}$ & 43.5-71.3 & 34.0\\
& $\phi_{new}^{contrast}/\phi_{new}^{contrast}$ & 72.5-86.3 & 58.5\\
\hline
\multirow{2}{5em}{$BT^2$ (+32)} & $\phi_{new}^{bt^2}/\phi_{old}$ & \textbf{53.6}-\textbf{74.5} & \textbf{37.5}\\
& $\phi_{new}^{bt^2}/\phi_{new}^{bt^2}$ & \textbf{76.9}-\textbf{88.2} & \textbf{70.4}\\
\bottomrule[1pt]
\end{tabular}}
\caption{Backward compatible experiments on Imagenet-500 to Imagenet-1k with both data change and model change. The old model uses Resnet50-128 architecure, while the new model uses a transformer ``ViT-B16'' \cite{DBLP:journals/corr/abs-2010-11929} pretrained on full training set of Imagenet21K.}
\label{table:Imagenet model change}
\end{center}
\vspace{-5mm}
\end{table}

\subsection{Modality Change}
\label{sec:modality change}

\noindent \textbf{Setting.} In addition to data and model change, we propose an even more challenging scenario where the modality changes. We consider the application of modality fusion, where a single gallery of image representations can support both image-to-image retrieval and text-to-image retrieval. In a standard setting, this needs to be done with two separate models: one trained for good image representations (with classification loss on a large-scale image datasets for example) and the other one trained to align the representations of images and text in the same representation space. The first model is usually good only for image-to-image retrieval, while the second model, although good for text-to-image matching, performs much worse than the first model in terms of image to image retrieval. One such popular model is CLIP~\cite{DBLP:journals/corr/abs-2103-00020}. CLIP comprises of a pair of text and image encoder, both of which are trained to align in representation. CLIP has been shown to be very effective for text-to-image matching but its performance on image-to-image matching lags behind. Denoting the image encoder of CLIP as $\phi^{clip-img}$ and referring to Table~\ref{table:Imagenetmodalitychange}, the CMC top 1 on Imagenet $\phi^{clip-img}/\phi^{clip-img}$ is only 54.7\% compared to using a specialized image-to-image retrieval model $\phi_{img}/\phi_{img}$ achieving 78.0\%, where $\phi_{img}$ is a pretrained Visual Transformer ``ViT-B16'' \cite{DBLP:journals/corr/abs-2010-11929} finetuned on Imagenet-1k classification. We believe that $BT^2$ has the potential to bridge this gap. Specifically, in the $BT^2$ setting, we use the text encoder of CLIP, denoted as $\phi^{clip-txt}$, as the old model and the ViT-B16 as $\phi_{new}'$.

We measure the performances of different baselines and our method by $\phi^{clip-text}/\phi_{img}$ and $\phi_{img}/\phi_{img}$, which are the performances of text-to-image and image-to-image retrievals respectively. For the former, we simulate text-to-image by using a pretrained GPT2 \cite{Radford2019LanguageMA} image captioning model to automatically generate captions for all images in Imagenet with ``vit-gpt2-image-captioning'' from \cite{DBLP:journals/corr/abs-1910-03771}. Subsequently, during evaluation, the queries are taken from the same set of captions (for which we know the corresponding images), then encoded ($\phi^{clip-txt}$) before the nearest image encodings are retrieved. Since we know the corresponding class of the image of $\phi^{clip-txt}$, we consider the retrieval to be correct if the retrieved image is from the same class. For comparison, we use $\phi^{clip-txt}/\phi^{clip-img}$ to denote the text-to-image retrieval performance with CLIP text model and CLIP image model.

\noindent \textbf{Results.} First, we notice that the performance of $\phi^{clip-text}/\phi^{clip-image}$, where we measure text-to-image retrieval performance with CLIP text model and CLIP image model, is relatively low compared to other models. This is mainly because we use automatic image captioning instead of manually annotated ones to caption the text descriptions associated with each image, so many of them are not accurate. Second, because of the challenge of modality change and noise in the training data, BCT fails to achieve backward compatibility (CMC top 1 on $\phi^{clip-text}/\phi_{image}^{bct}$ is 9.1\% compared to 11.7\% on $\phi^{clip-txt}/\phi^{clip-img}$) and the performance on $\phi_{img}^{bct}/\phi_{img}^{bct}$ is also significantly hurt by 4.4\% (CMC top 1 of 73.7\% compared to 78.0\%). On the other hand, although Contrast achieves backward compatibility, its image-to-image retrieval performance is extremely unreliable (CMC top 1 of 48.0\% compared to 78.0\%). Finally, we observe that $BT^2$ is particularly robust in this setting with rigorous backward compatibility (CMC top 1 11.4\% compared to 11.7\%) and marginal loss in $\phi_{img}^{bt^2}/\phi_{img}^{bt^2}$ (CMC top 1 77.6\% compared to 78.0\%). We also want to note that unlike the experiments of data change and model change, we set the dimension of $\phi_{img}^{bct}$ and $\phi^{clip-txt}$ to be 512, same as the dimension used in CLIP while still managing to keep the dimension expansion of $\phi_{img}^{bt^2}$ to 32. It shows that our $BT^2$ is extremely ``dimension efficient'' in the sense that we only need to expand the dimension by 6.25\%.

\begin{table}
\begin{center}
\resizebox{\columnwidth}{!}{%
\begin{tabular}{ c c c c }
\toprule[1pt]
\multirow{2}{5em}{Method} & \multirow{2}{3em}{Case} & CMC & \multirow{2}{5em}{mAP@1.0} \\
& & top1-top5 &\\
\hline
\multirow{4}{5em}{Independent} & $\phi^{clip-txt}/\phi^{clip-img}$ & 11.7-25.0 & 7.0\\
& $\phi^{clip-txt}/\phi_{img}$ & 0.1-0.4 & 0.3\\
&$\phi^{clip-img}/\phi^{clip-img}$ & 54.7-78.0 & 21.8\\
& $\phi_{img}/\phi_{img}$ & 78.0-87.5 & 72.4\\
\hline
\multirow{2}{5em}{BCT} & $\phi^{clip-txt}/\phi_{img}^{bct}$  & 9.1-21.4 & 12.1\\
& $\phi_{img}^{bct}/\phi_{img}^{bct}$ & 73.7-85.9 & 65.9\\
\hline
\multirow{2}{5em}{BCT (+32)} & $\phi^{clip-txt}/\phi_{img}^{bct(+32)}$ & 9.2-21.8 & 12.1\\
& $\phi_{img}^{bct(+32)}/\phi_{img}^{bct(+32)}$ & 73.8-85.5 & 65.5\\
\hline
\multirow{2}{5em}{Contrast} & $\phi^{clip-txt}/\phi_{img}^{contrast}$ & \textbf{13.0}-\textbf{30.0} & 8.8\\
& $\phi_{img}^{contrast}/\phi_{img}^{contrast}$ & 48.0-68.2 & 23.1\\
\hline
\multirow{2}{5em}{$BT^2$ (Ours)} & $\phi^{clip-text}/\phi_{img}^{bt^2}$ & 11.4-25.6 & \textbf{13.6}\\
& $\phi_{img}^{bt^2}/\phi_{img}^{bt^2}$ & \textbf{77.6}-\textbf{87.4} & \textbf{71.5}\\
\bottomrule[1pt]
\end{tabular}}
\caption{Backward compatible experiments on Imagenet-1k with modality change. The old model uses a pretrained CLIP text encoder with automatically generated text, while the new model uses a transformer pretrained on full training set of Imagenet21K.}
\label{table:Imagenetmodalitychange}
\end{center}
\vspace{-5mm}
\end{table}

\subsection{Series of Model Updates}

The evolution of deep learning brought in an era of very prolific scientific production, bringing us new and better model designs and architectures every so often. From the early days of the past decade (2010-2022) when AlexNet \cite{NIPS2012_c399862d} was first introduced, we have seen an evolution that subsequently brought us VGGNet13 \cite{simonyan2014very}, ResNet50 \cite{DBLP:journals/corr/HeZRS15}, and finally ViT \cite{DBLP:journals/corr/abs-2010-11929}. There are of course numerous other model designs but we will focus on this list of better-known architectures. The experiments in this section is as follow. We first train a VGGNet13, $\phi_{vgg}$, that is backward compatible with AlexNet. Then, we train a ResNet50, $\phi_{res}$, model that is backward compatible with $\phi_{vgg}$, and finally we train a ViT, $\phi_{vit}$, that is compatible with $\phi_{res}$. Therefore, $\phi_{vit}$ would be a model that is backward compatible with all the earlier models. All models are trained on Imagenet-1k training and results are tested on Imagenet-1k validation set. Our $BT^2$ adds 32 extra dimensions for each update, while all other models use an embedding size of 128.

\begin{table}
\begin{center}
\resizebox{\columnwidth}{!}{%
\begin{tabular}{ c c c c }
\toprule[1pt]
\multirow{2}{5em}{Method} & \multirow{2}{3em}{Case} & CMC & \multirow{2}{5em}{mAP@1.0} \\
& & top1-top5 &\\
\hline
\multirow{4}{5em}{Independent} & $\phi_{alex}/\phi_{alex}$ & 46.6-66.3 & 29.1\\
& $\phi_{vgg}/\phi_{vgg}$ & 63.2-79.0 & 49.6\\
& $\phi_{res}/\phi_{res}$ & 67.9-81.4 & 52.3\\
& $\phi_{vit}/\phi_{vit}$ & 78.0-87.5 & 72.4\\
\hline
\multirow{9}{5em}{BCT} & $\phi_{vgg}^{bct}/\phi_{alex}$  & 54.4-74.1 & 36.2\\
& $\phi_{vgg}^{bct}/\phi_{vgg}^{bct}$ & 58.4-75.4 & 47.0\\
& $\phi_{res}^{bct}/\phi_{alex}^{bct}$ & 46.0-71.9 & 30.6\\
& $\phi_{res}^{bct}/\phi_{vgg}^{bct}$ & 48.9-75.2 & 44.4\\
& $\phi_{res}^{bct}/\phi_{res}^{bct}$ & 64.3-79.1 & 52.7\\
& $\phi_{vit}^{bct}/\phi_{alex}^{bct}$ & 54.9-82.0 & 36.3\\
& $\phi_{vit}^{bct}/\phi_{vgg}^{bct}$ & 57.5-84.1 & 50.5\\
& $\phi_{vit}^{bct}/\phi_{res}^{bct}$ & 70.3-85.1 & 57.0\\
& $\phi_{vit}^{bct}/\phi_{vit}^{bct}$ & 73.9-86.0 & 65.8\\
\hline
\multirow{9}{5em}{$BT^2$ (ours)} & $\phi_{vgg}^{bt^2}/\phi_{alex}$  & \textbf{56.5}-\textbf{75.6} & \textbf{37.1}\\
& $\phi_{vgg}^{bt^2}/\phi_{vgg }^{bt^2}$ & \textbf{61.0}-\textbf{77.2} & \textbf{48.5}\\
& $\phi_{res}^{bt^2}/\phi_{alex}^{bt^2}$ & \textbf{56.7}-\textbf{78.5} & \textbf{37.2}\\
& $\phi_{res}^{bt^2}/\phi_{vgg}^{bt^2}$ & \textbf{61.5}-\textbf{80.8} & \textbf{50.6}\\
& $\phi_{res}^{bt^2}/\phi_{res}^{bt^2}$ & \textbf{66.6}-\textbf{80.8} & \textbf{56.8}\\
& $\phi_{vit}^{bt^2}/\phi_{alex}^{bt^2}$ & \textbf{57.9}-\textbf{83.5} & \textbf{37.6}\\
& $\phi_{vit}^{bt^2}/\phi_{vgg}^{bt^2}$ & \textbf{62.5}-\textbf{86.5} & \textbf{52.7}\\
& $\phi_{vit}^{bt^2}/\phi_{res}^{bt^2}$ & \textbf{72.0}-\textbf{87.0} & \textbf{60.6}\\
& $\phi_{vit}^{bt^2}/\phi_{vit}^{bt^2}$ & \textbf{75.6}-\textbf{87.4} & \textbf{68.0}\\
\bottomrule[1pt]
\end{tabular}}
\caption{Backward compatible experiments on a series of model updates on Imagenet-1k. Our $BT^2$ adds 32 extra dimensions for each update, all other models use an embedding size of 128.}
\label{table: a series of updates}
\end{center}
\vspace{-8mm}
\end{table}

\noindent \textbf{Results.} The results are shown in Table \ref{table: a series of updates}. Note that because BCT (+32) and Contrast did not show clear advantage over BCT in previous experiments, we only compare our method with BCT here. We first note the failure of BCT in this challenging setting. In the first update from AlexNet to VGGNet13, although backward compatiblility is achieved (for Top1, $\phi_{vgg}^{bct}/\phi_{alex}$ is 54.4\% while independent $\phi_{alex}/\phi_{alex}$ is 46.6\%), the performance of $\phi_{vgg}^{bct}/\phi_{vgg}^{bct}$ is significantly hurt (for Top1, $\phi_{vgg}^{bct}/\phi_{vgg}^{bct}$ is 58.4\% while independent $\phi_{vgg}/\phi_{vgg}$ is 63.2\%). Furthermore, after another round of update in model architecture from VGGNet13 to ResNet50, the model does not maintain a decent backward compatibility with its former versions $\phi_{alex}$ and $\phi^{bct}_{vgg}$ with a loss of 0.6\% on $\phi_{res}^{bct}/\phi_{alex}^{bct}$ and 4.3\% on $\phi_{res}^{bct}/\phi_{vgg}^{bct}$ compared to independent $\phi_{alex}/\phi_{alex}$ and $\phi_{vgg}/\phi_{vgg}$ in Top1. Although its backward compatibility is improved after updating to $\phi^{bct}_{vit}$ possibly because of the power of pretraining for ViT, it still hurts the performance of the new model significantly (for Top1, $\phi^{bct}_{vit}/\phi^{bct}_{vit}$ is only 73.9\% compared to independent $\phi_{vit}/\phi_{vit}$ with 78.0\%). On the contrary, we observe that by a strategic use of extra dimensions, our $BT^2$ maintains backward compatibility at each stage of the sequential updates and closes the gap between the performance of the independent new models and the models trained with BCT. In summary, $BT^2$ shows a clear advantage on all rows across the board with gains up to 12.6\%.

\section{Ablations on Extra Dimensions}
\begin{table}
\begin{center}
\resizebox{\columnwidth}{!}{%
\begin{tabular}{ c c c c }
\toprule[1pt]
\multirow{2}{5em}{Method} & \multirow{2}{3em}{Case} & CMC & \multirow{2}{5em}{mAP@1.0} \\
& & top1-top5 &\\
\hline
\multirow{3}{5em}{Independent} & $\phi_{old}/\phi_{old}$ & 43.1-58.3 & 30.9\\
& $\phi_{new}'/\phi_{old}$ & 0.0-0.2 & 0.1\\
& $\phi_{new}'/\phi_{new}'$ & 78.0-87.5 & 72.4\\
\hline
\multirow{2}{5em}{BCT} & $\phi_{new}^{bct}/\phi_{old}$  & 41.1-69.5 & 33.0\\
& $\phi_{new}^{bct}/\phi_{new}^{bct}$ & 74.8-86.7 & 66.5\\
\hline
 \multirow{2}{5em}{$BT^2$ (+8)} & $\phi_{new}^{+8}/\phi_{old}$ & 50.7-75.9 & 37.1\\
& $\phi_{new}^{+8}/\phi_{new}^{+8}$ & 74.8-87.0 & 66.0\\
\hline
\multirow{2}{5em}{$BT^2$ (+16)} & $\phi_{new}^{+16}/\phi_{old}$ & 51.6-\textbf{76.1} & 37.6\\
& $\phi_{new}^{+16}/\phi_{new}^{+16}$ & 76.4-88.1 & 69.0\\
\hline
\multirow{2}{5em}{$BT^2$ (+32)} & $\phi_{new}^{+32}/\phi_{old}$ & \textbf{53.6}-74.5 & 37.5\\
& $\phi_{new}^{+32}/\phi_{new}^{+32}$ & 76.9-88.2 & 70.4\\
\hline
\multirow{2}{5em}{$BT^2$ (+128)} & $\phi_{new}^{+128}/\phi_{old}$ & 53.4-75.0 & \textbf{37.9}\\
& $\phi_{new}^{+128}/\phi_{new}^{+128}$ & \textbf{77.8}-\textbf{88.5} & \textbf{71.1}\\
\bottomrule[1pt]
\end{tabular}}
\caption{Dimension ablation experiments on Imagenet-500 to Imagenet-1k with both data change and model change. The old model uses Resnet50-128 architecure, while the new model uses a transformer pretrained on full training set of Imagenet21K. Our $BT^2$ adds 32 extra dimensions for each update, all other models use an embedding size of 128.}
\label{table:dimension ablations}
\end{center}
\vspace{-5mm}
\end{table}

We now answer the question on how many additional dimensions are needed for $BT^2$. We carry out ablation experiments in the setting of ResNet50 to ViT (with both data change and model change) on Imagenet-1k. All settings are the same as in Section \ref{sec:model change} except we vary the number of extra dimensions in our $BT^2$. Results of the ablations are shown in Table \ref{table:dimension ablations}, where we compare BCT (no extra dimension) with $BT^2$ extra 8, 16, 32, 128 dimensions. We observe that $BT^2$ already shows a clear advantage over BCT with as few as 8 extra dimensions, showing the effectiveness of our proposed Basis Transformation Block. As extra dimensions grow from 8 to 32, though less significant, we observe an overall trend of gradual improvement both in terms of $\phi_{new}^{+n}/\phi_{new}^{+n}$ and $\phi_{new}^{+n}/\phi_{old}$. As a result, the CMC top 1 of $BT^2$ (+32) is 2.9\% and 2.1\% higher than $BT^2$ (+8) in terms of $\phi_{new}^{+n}/\phi_{new}^{+n}$ and $\phi_{new}^{+n}/\phi_{old}$ respectively. Finally, we notice that the improvement from $BT^2$ (+32) to $BT^2$ (+128) is somewhat marginal by -0.2\% and 0.9\% in terms of $\phi_{new}^{+n}/\phi_{new}^{+n}$ and $\phi_{new}^{+n}/\phi_{old}$ respectively. This shows that some of the information of $\phi_{new}'$ and $\phi_{old}$ can be shared so that we do not need as much as +128 to capture extra information from $\phi_{new}'$.
\section{Conclusions}
We presented $BT^2$ in this paper, a method for backward compatibility that makes use of additional dimensions \emph{efficiently}. In spite of this, one of the open questions following this work is that the size of the representation will still grow over time especially after multiple model updates. Eventually, system practitioners will have to fully backfill the gallery to reset. However, we hope that $BT^2$ will ``buy enough time'' to backfill real world galleries with $\phi_{new}'$ that usually can contain millions of samples.  


{\small
\bibliographystyle{ieee_fullname}
\bibliography{egbib}
}
\newpage
\appendix

\section{Techniques from Representation Learning}
The task of Backward Compatible Representation Learning exploits techniques from the field of representation learning \cite{8395024, 6472238, kaya2019deep, bellet2015metric, hoffer2015deep, jaiswal2020survey}, where classification \cite{DBLP:journals/corr/LiuWYLRS17, DBLP:journals/corr/WangXCY17, DBLP:journals/corr/abs-1811-12649, DBLP:journals/corr/abs-1801-05599, DBLP:journals/corr/abs-1801-09414}, metric learning \cite{DBLP:journals/corr/abs-2003-08505, https://doi.org/10.48550/arxiv.1904.06627, DBLP:journals/corr/WuMSK17, 8953619}, and contrastive learning \cite{DBLP:journals/corr/abs-2002-05709, He_2020_CVPR, NEURIPS2020_f3ada80d} are some major methods. For simplicity and better alignment with previous works in backward compatible representation learning \cite{8953998, DBLP:journals/corr/abs-2108-01958, https://doi.org/10.48550/arxiv.2203.01583, ramanujan2022forward}, we adopt the classification loss for training the representation model.

\section{Cosine Similarity vs Euclidean Distance}
We note that some of the previous works on representation learning and backward compatibility use the Euclidean Distance for retrieval \cite{9156906, ramanujan2022forward} while others use Cosine Similarity \cite{DBLP:journals/corr/abs-2108-01958, DBLP:journals/corr/abs-1904-06627, DBLP:journals/corr/abs-2006-07733, DBLP:journals/corr/abs-2003-08505, DBLP:journals/corr/abs-2103-00020}. Preliminary experiments that we conducted did not find 
any clear superiority between the metrics when compared with public results in \cite{9156906, ramanujan2022forward}. We adopt cosine similarity which provides for clearer analysis and better compatibility with our experiments on multi-modality.

\section{Proof of Lemmas}
For completeness, we provide proofs to the lemmas stated in the main text.
\begin{proof}[Proof of Lemma~\pref{lemma:counterexample}] 
We define an image space $\mathcal{X}$, an n-dimensional representation space $\mathbb{R}^n$, and two representation functions $\phi_{old}, \phi_{new}: \mathcal{X} -> \mathbb{R}^n$ that maps images to a unit ball in $\mathbb{R}^n$. We consider the distance metric $d$ being the negative cosine similarity, and $\forall x, \|\phi_{old}(x)\|_2 = \|\phi_{new}(x)\|_2 = 1$

To construct this counterexample, for two images in the gallery, $x_1$ and $x_2$  of the same class $y$ where the representations of $x_1, x_2$ are $\phi_{old}(x_1),\phi_{old}(x_1)$. The third image as a query, $\bar{x}$ of the same class $y$, has its old representation and new representation as $\phi_{old}(\Bar{x}), \phi_{new}(\Bar{x})$. We consider a specific case where $\phi_{old}(\Bar{x})$ is close to the cone spanned by $\phi_{old}(x_1), \phi_{old}(x_2)$ defined by by:
\begin{align*}
    \phi_{old}(\Bar{x}) = a\phi_{old}(x_1) + b \phi_{old}(x_2) + \epsilon,\\
    \text{for some } a, b > 0, \text{ and small } \epsilon
\end{align*}
Let the projection of $\phi_{old}(\Bar{x})$ to the plane of $\phi_{old}(x_1), \phi_{old}(x_2)$ be $P(\phi_{old}(\Bar{x}))$. Let the angle between $P(\phi_{old}(\Bar{x}))$ and $\phi_{old}(x_1), \phi_{old}(x_2)$ be $\theta_1, \theta_2$, and let the angle between $P(\phi_{old}(\Bar{x}))$ and $\phi_{old}(\Bar{x})$ be $\delta \theta$ so that $\sin{\delta \theta} = \epsilon$. Similarly, let the projection of $\phi_{new}(\Bar{x})$ be $P(\phi_{new}(\Bar{x}))$, whose angle with $\phi_{old}(x_1), \phi_{old}(x_2)$ be $\theta_3, \theta_4$, and its angle with $\phi_{new}(\Bar{x})$ be $\delta \theta'$. $\theta_1, \theta_2, \theta_3, \theta_4 \in [0, \pi], \delta \theta, \delta \theta' \in [0, \frac{\pi}{2}]$.

By the criterion for backward compatibility defined in Definition \ref{backward compatibility}, we have:
\begin{align*}
    d(\phi_{new}(\Bar{x}), \phi_{old}(x_1)) \le d(\phi_{old}(\Bar{x}), \phi_{old}(x_1))\\
    d(\phi_{new}(\Bar{x}), \phi_{old}(x_2)) \le d(\phi_{old}(\Bar{x}), \phi_{old}(x_2)), \stepcounter{equation}\tag{\theequation}\label{equation:1}
\end{align*}
which gives us
\begin{align*}
    \cos \theta_3 \cos \delta \theta' \ge \cos \theta_1 \cos \delta \theta\\
    \cos \theta_4 \cos \delta \theta' \ge \cos \theta_2 \cos \delta \theta
\end{align*}
To bound $\delta \theta'$, we first notice that $\theta_1 + \theta_2 = \theta(\phi_{old}(x_1), \phi_{old}(x_2)) \le \pi$, with $\theta(\phi_{old}(x_1), \phi_{old}(x_2))$ being the angle between $\phi_{old}(x_1), \phi_{old}(x_2)$, because $\phi_{old}(\Bar{x})$ lies in the cone. Because of the constraint in Equation \ref{equation:1}, $\phi_{new}(\Bar{x})$ must also lie in the cone. Therefore, $\theta_1 + \theta_2 = \theta_3 + \theta_4 = \theta(\phi_{old}(x_1), \phi_{old}(x_2)) \le \pi$, which yields
\begin{align*}
    (\theta_1 - \theta_3)(\theta_2 - \theta_4) \le 0\\
    (\cos \theta_1 - \cos \theta_3)(\cos \theta_2 - \cos \theta_4) \le 0.\\ \stepcounter{equation}\tag{\theequation}\label{equation:2}
\end{align*}
Comparing Equation \ref{equation:1} and Equation \ref{equation:2}, we conclude that $\delta \theta' \le \delta \theta$, so that $\cos \delta \theta' \ge \cos \delta \theta$.

To further bound $\cos (\theta_1 - \theta_3)$, by inspecting Equation \ref{equation:1}, in the case of $\theta_1<\theta_3$ we have:
\begin{align*}
    \cos \theta_3 \cos \delta \theta' \ge & \cos \theta_1 \cos \delta \theta\\
    \cos \theta_3 \ge & \cos \theta_1 \frac{\cos \delta \theta}{\cos \delta \theta'}\\
    \cos \theta_3 \ge & \cos \theta_1 \cos \delta \theta\\
    \cos \theta_3 \ge & \cos \theta_1 \sqrt{1-\epsilon^2}\\
    \cos \theta_1 - \cos \theta_3 \le & \frac{1 - \sqrt{1 - \epsilon^2}}{\sqrt{1 - \epsilon^2}},\stepcounter{equation}\tag{\theequation}\label{equation:3}
\end{align*}
where the third inequality follows by upperbounding $\cos \delta \theta'$ to be 1, the fourth inequality by substituting $\delta \theta$ with $\epsilon$, the fifth inequality follows by upperbounding $\cos \theta_3$ to be 1. By further expanding Equation \ref{equation:3}, 
\begin{align*}
    \cos \theta_1 - \cos \theta_3 \le & \frac{1 - \sqrt{1 - \epsilon^2}}{\sqrt{1 - \epsilon^2}}\\
    2 \sin \frac{\theta_1+\theta_3}{2} \sin \frac{\theta_3 - \theta_1}{2} \le & \frac{1 - \sqrt{1 - \epsilon^2}}{\sqrt{1 - \epsilon^2}}\\
    \sin^2 \frac{\theta_3 - \theta_1}{2} \le & \frac{1 - \sqrt{1 - \epsilon^2}}{2\sqrt{1 - \epsilon^2}}\\
    \cos^2 \frac{\theta_3 - \theta_1}{2} \ge & 1 - \frac{1 - \sqrt{1 - \epsilon^2}}{2\sqrt{1 - \epsilon^2}}\\
    \cos (\theta_3 - \theta_1) \ge & 1 - \frac{1 - \sqrt{1 - \epsilon^2}}{\sqrt{1 - \epsilon^2}},\stepcounter{equation}\tag{\theequation}\label{equation:4}
\end{align*}
where the third inequality follows from lowerbounding $\sin \frac{\theta_1 + \theta_3}{2}$ by $\sin \frac{\theta_3 - \theta_1}{2}$.

Similarly, in the case of $\theta_1 \ge \theta_3$, we have $\theta_2 \le \theta_4$, so that $\cos (\theta_3 - \theta_1) = \cos (\theta_4 - \theta_2) \ge  1 - \frac{1 - \sqrt{1 - \epsilon^2}}{\sqrt{1 - \epsilon^2}}$. Therefore, we have in all cases, $\cos (\theta_3 - \theta_1) \ge  1 - \frac{1 - \sqrt{1 - \epsilon^2}}{\sqrt{1 - \epsilon^2}}$.

With both $\cos (\theta_3 - \theta_1)$ and $\cos \delta \theta'$, we have the cosine similarity between $\phi_{old}(\Bar{x})$ and $\phi_{new}(\Bar{x})$, $\cos (\phi_{old}(\Bar{x}), \phi_{new}(\Bar{x}))$ being bounded by
\begin{align*}
    &\cos (\phi_{old}(\Bar{x}), \phi_{new}(\Bar{x}))\\ 
    \ge& \cos (\phi_{old}(\Bar{x}), P(\phi_{old}(\Bar{x})))  \cos (\phi_{new}(\Bar{x}), P(\phi_{old}(\Bar{x})))\\
    \ge& \sqrt{1 - \epsilon^2} \cos (P(\phi_{new}(\Bar{x})), \phi_{new}(\Bar{x}))   \\
    & \times \cos (P(\phi_{new}(\Bar{x})), P(\phi_{old}(\Bar{x})))\\
    \ge& (1 - \epsilon^2)(1 - \frac{1 - \sqrt{1 - \epsilon^2}}{\sqrt{1 - \epsilon^2}})
\end{align*}

Therefore, we show that in order to be backward compatible with $\phi_{old}(x_1), \phi_{old}(x_2)$, $\phi_{new}(\Bar{x})$ is restricted within a small angle from $\phi_{old}(\Bar{x})$, with $\cos (\phi_{old}(\Bar{x}), \phi_{new}(\Bar{x})) \ge (1 - \epsilon^2)(1 - \frac{1 - \sqrt{1 - \epsilon^2}}{\sqrt{1 - \epsilon^2}})$. This limits the room of improvement of $\phi_{new}(\Bar{x})$ over $\phi_{old}(\Bar{x})$, especially when $\phi_{old}(\Bar{x})$ is not good.

Proof of Lemma \ref{lemma:matrix exponentiation} can be found in \cite{geometric}.
\begin{proof}[Proof of Lemma~\pref{lemma:change of basis}] 
For any orthonormal matrix $P$, and representation function $\phi$, any images $x_1, x_2$, we have
\begin{align*}
    &(P(\phi(x_1)))^\top P(\phi(x_2))\\
    =&\phi(x_1)^\top P^\top P \phi(x_2)\\
    =&\phi(x_1)^\top (P^\top P) \phi(x_2)\\
    =&\phi(x_1)^\top \phi(x_2)
\end{align*}

\end{proof}

\section{Sample Captions for Imagenet-1k}
\label{app:sample_captions}
We did not find existing dataset that simultaneously supports both evaluation of image-to-image retrieval representations and image-to-text representations. To our purpose of modality fusion, we generate automatic captions for Imagenet-1k with ``vit-gpt2-image-captioning'' from \cite{DBLP:journals/corr/abs-1910-03771}. We provide sample captions generated in Figure \ref{fig:sample_captions}. We observe that although automatic image captions capture daily pictures like dogs and benches well, it does not recognize other less common pictures like wild animals and pills. This is an expected behavior because automatic image captioning models might have encountered more daily pictures during the training than less common ones. Learning under such strong noise pose a significant challenge to the robustness of different methods, and it also causes the evaluation of text-to-image retrieval accuracy lower than it should be.

\begin{figure}[!ht]
  \centering
   \includegraphics[width=\linewidth]{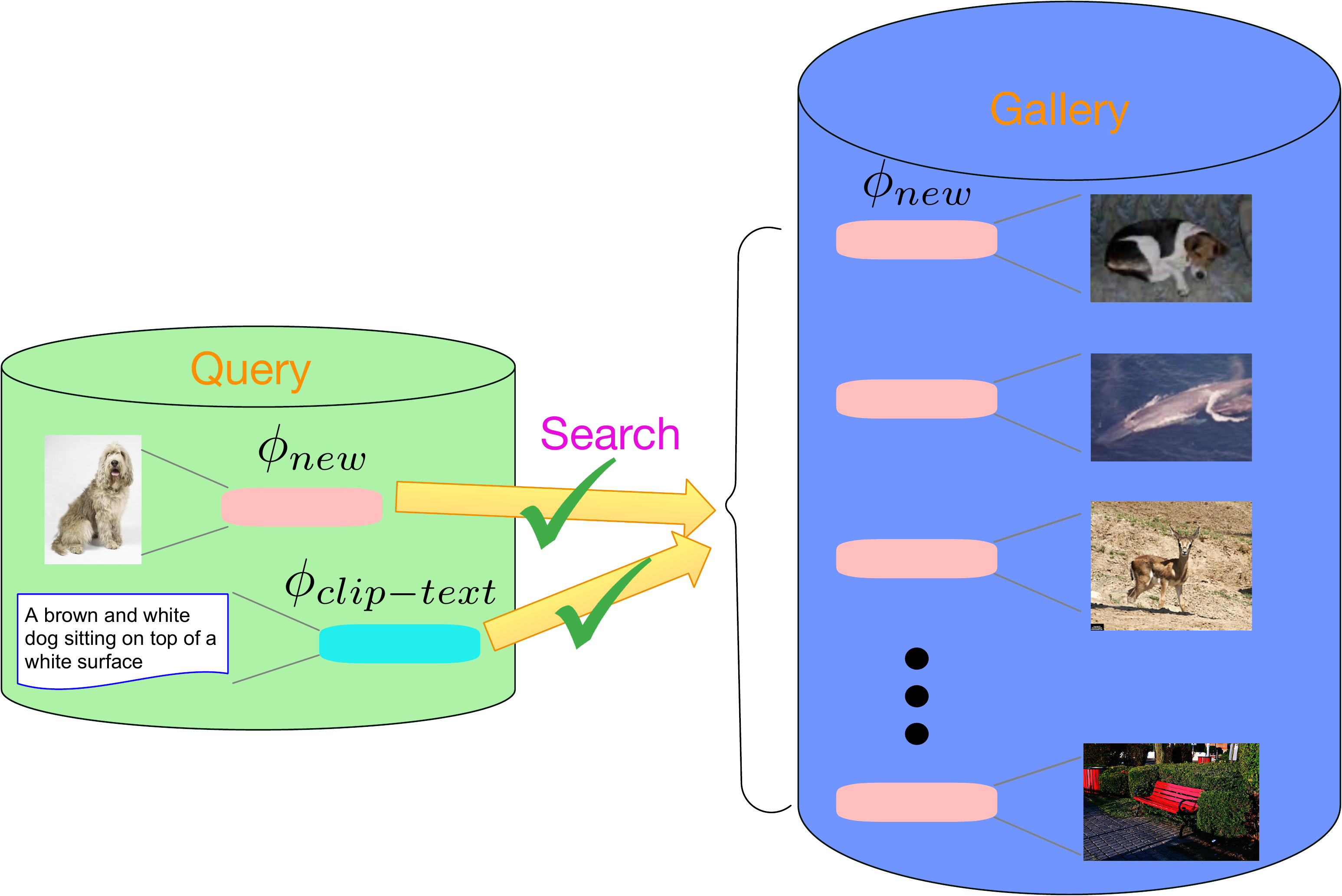}
   \caption{An illustration of the idea of modality fusion. A gallery of images is encoded with a single representation $\phi_{new}$ but can support query with images encoded by $\phi_{new}$ and text encoded by $\phi_{clip-text}$ at the same time.}
   \label{fig:modality_fusion}
\end{figure}

\begin{figure*}[!ht]
  \centering
   \includegraphics[width=\linewidth]{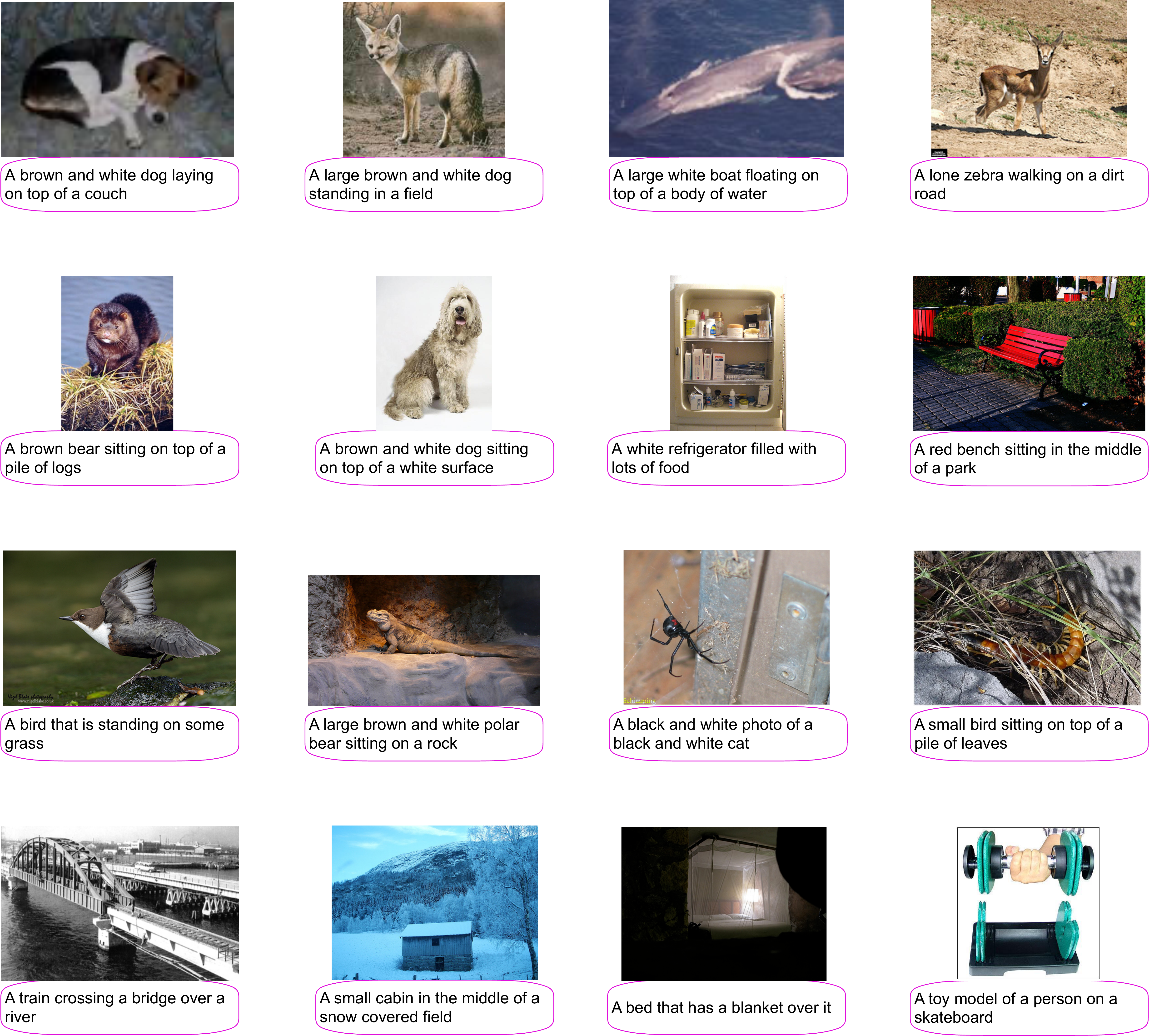}
   \caption{Sample captions automatically generated for Imagenet-1k with ``vit-gpt2-image-captioning'' from \cite{DBLP:journals/corr/abs-1910-03771}.}
   \label{fig:sample_captions}
\end{figure*}


\end{proof}

\section{Confidence Intervals}
Because of limited computational resources, we are unable to provide confidence intervals for all of our experiments. To get a sense of the variances of the experiments, we conduct backward compatible experiments on a subset of Imagenet-1k with 50k images (50 images from each class). We use ResNet50-128 model architecture for both the old model and the new model. Old models are trained using 500 classes of our constructed Imagenet-1k subset while the new models have access to the entire 1000 classes. The independent models ($\phi_{old}$ and $\phi_{new}'$) are only trained once, but we calculate means and standard deviations over 5 random seeds of training the new model.

As shown in Table \ref{Table:confidence_intervals}, we found that the standard deviations for both BCT and $BT^2$ are relatively small with respect to all the metrics (below $0.5\%$), and the advantage of $BT^2$ over $BCT$ is indeed statistically significant in both $\phi_{new}/\phi_{new}$ and $\phi_{new}/\phi_{old}$. For example, in terms of $\phi_{new}/\phi_{new}$ Top-1 accuracy, $BT^2$ achieves 21.4\% while BCT achieves 18.3\%. This gain of 3.1\% is statistically significant considering the standard deviations of the results are only 0.3\% and 0.4\% respectively. We hope this supplementary experiment can provide a rough idea of the degree of randomness in our backward compatible experiments.

\begin{table}[h!]
 \resizebox{\columnwidth}{!}{%
\begin{tabular}{lccc}\\\toprule  
 Method& Case & Top1-Top5 & mAP \\\midrule
 \multirow{2}{2em}{Independent} & $\phi_{old}/\phi_{old}$ & $10.3$-$25.0$ & $6.2$\\ 
& $\phi'_{new}/\phi'_{new}$ & $17.8$-$37.5$ & $10.5$\\  \midrule
\multirow{2}{2em}{BCT} & $\phi^{bct}_{new}/\phi^{bct}_{old}$ & $11.5 \pm 0.1$-$29.3 \pm 0.3$ & $7.6 \pm 0.1$\\ 
& $\phi^{bct}_{new}/\phi^{bct}_{new}$ & $18.3 \pm 0.4$-$38.7 \pm 0.5$ & $12.7 \pm 0.1$\\  \midrule
\multirow{2}{2em}{$BT^2$} & $\phi^{bt^2}_{new}/\phi^{bt^2}_{old}$ & \textbf{$12.6 \pm 0.2$}-\textbf{$31.0 \pm 0.3$} & \textbf{$8.0 \pm 0.0$}\\  
& $\phi^{bt^2}_{new}/\phi^{bt^2}_{new}$ & \textbf{$21.4 \pm 0.3$}-\textbf{$42.6 \pm 0.3$} &\textbf{$14.6 \pm 0.1$} \\  \bottomrule
\end{tabular}}
\caption{Backward compatible experiments on Imagenet-500 to Imagenet-1k (a 50k images subset) with only data change. Both the old model and the new model uses Resnet50-128 architecure.}\label{Table:confidence_intervals}
\end{table} 

\end{document}